\documentclass{article}

\PassOptionsToPackage{numbers}{natbib}

\usepackage{microtype}
\usepackage{graphicx}
\usepackage{booktabs}

\PassOptionsToPackage{hyphens}{url}\usepackage{hyperref}

\usepackage[accepted]{icml2020}

\usepackage[utf8]{inputenc}
\usepackage[T1]{fontenc}
\usepackage{amsfonts}
\usepackage{amsmath}
\usepackage{amsthm}
\usepackage{bm}
\usepackage{dsfont}
\usepackage{url}
\usepackage{nicefrac}
\usepackage{enumitem}
\usepackage{caption}
\usepackage{subcaption}
\usepackage[toc,page]{appendix}

\usepackage{listings}
\usepackage{mdframed}
\mdfsetup{skipbelow=4pt,skipabove=6pt,leftmargin=4pt,rightmargin=4pt,align=left,usetwoside=false}
\mdfdefinestyle{listingstyle}{
  backgroundcolor=gray!3,
  linewidth=2pt,linecolor=gray!30,
  outerlinewidth=5pt,outerlinecolor=black,
  rightline=false,topline=false,bottomline=false,
  innerleftmargin=6pt,innerrightmargin=3pt,innertopmargin=0pt,innerbottommargin=0pt,
}
\surroundwithmdframed[style=listingstyle]{lstlisting}

\DeclareMathOperator{\mass}{mass}
\DeclareMathOperator{\unsampledFraction}{unsampledFraction}
\DeclareMathOperator{\edgeProbability}{edgeProbability}
\DeclareMathOperator{\gumbel}{Gumbel}
\DeclareMathOperator{\costDelta}{costDelta}
\DeclareMathOperator{\children}{children}

\usepackage{algorithm}
\usepackage[noend]{algpseudocode}
\usepackage{algorithmicx}
\usepackage{float}
\newfloat{algorithm}{t}{}

\makeatletter
\algnewcommand{\LineComment}[1]{\Statex \hskip\ALG@thistlm \(\triangleright\) #1}
\makeatother

\usepackage{etoolbox}
\usepackage{tikz}
\usetikzlibrary{tikzmark}
\usetikzlibrary{calc}

\newcommand{\ALGtikzmarkcolor}{lightgray}
\newcommand{\ALGtikzmarkextraindent}{4pt}
\newcommand{\ALGtikzmarkverticaloffsetstart}{-.7ex}
\newcommand{\ALGtikzmarkverticaloffsetend}{-.5ex}
\makeatletter
\newcounter{ALG@tikzmark@tempcnta}

\newcommand\ALG@tikzmark@start{%
    \global\let\ALG@tikzmark@last\ALG@tikzmark@starttext%
    \expandafter\edef\csname ALG@tikzmark@\theALG@nested\endcsname{\theALG@tikzmark@tempcnta}%
    \tikzmark{ALG@tikzmark@start@\csname ALG@tikzmark@\theALG@nested\endcsname}%
    \addtocounter{ALG@tikzmark@tempcnta}{1}%
}

\def\ALG@tikzmark@starttext{start}
\newcommand\ALG@tikzmark@end{%
    \ifx\ALG@tikzmark@last\ALG@tikzmark@starttext
    \else
        \tikzmark{ALG@tikzmark@end@\csname ALG@tikzmark@\theALG@nested\endcsname}%
        \tikz[overlay,remember picture] \draw[\ALGtikzmarkcolor] let \p{S}=($(pic cs:ALG@tikzmark@start@\csname ALG@tikzmark@\theALG@nested\endcsname)+(\ALGtikzmarkextraindent,\ALGtikzmarkverticaloffsetstart)$), \p{E}=($(pic cs:ALG@tikzmark@end@\csname ALG@tikzmark@\theALG@nested\endcsname)+(\ALGtikzmarkextraindent,\ALGtikzmarkverticaloffsetend)$) in (\x{S},\y{S})--(\x{S},\y{E});%
    \fi
    \gdef\ALG@tikzmark@last{end}%
}

\apptocmd{\ALG@beginblock}{\ALG@tikzmark@start}{}{\errmessage{failed to patch}}
\pretocmd{\ALG@endblock}{\ALG@tikzmark@end}{}{\errmessage{failed to patch}}
\makeatother

\newcommand{\code}[1]{\texttt{#1}}
\newcommand{\calP}{\mathcal{P}}
\newcommand{\calC}{\mathcal{C}}

\newcommand{\calF}{\mathcal{F}}
\newcommand{\WOR}{\textsc{\tiny WOR}}

\newtheorem{theorem}{Theorem}

\begin{document}

\twocolumn[
\icmltitle{Incremental Sampling Without Replacement for Sequence Models}



\icmlsetsymbol{equal}{*}

\begin{icmlauthorlist}
\icmlauthor{Kensen Shi}{google}
\icmlauthor{David Bieber}{google}
\icmlauthor{Charles Sutton}{google}
\end{icmlauthorlist}

\icmlaffiliation{google}{Google}
\icmlcorrespondingauthor{Kensen Shi}{kshi@google.com}

\icmlkeywords{sampling, without replacement, sequence models, program synthesis, combinatorial optimization}

\vskip 0.3in
]

\printAffiliationsAndNotice{}  

\begin{abstract}
    Sampling is a fundamental technique, and sampling \emph{without replacement} is often desirable when duplicate samples are not beneficial. Within machine learning, sampling is useful for generating diverse outputs from a trained model. We present an elegant procedure for sampling without replacement from a broad class of randomized programs, including generative neural models that construct outputs sequentially. Our procedure is efficient even for exponentially-large output spaces. Unlike prior work, our approach is \emph{incremental}, i.e., samples can be drawn one at a time, allowing for increased flexibility. We also present a new estimator for computing expectations from samples drawn without replacement. We show that incremental sampling without replacement is applicable to many domains, e.g., program synthesis and combinatorial optimization.
\end{abstract}

\section{Introduction}

Sampling from programmatically-defined distributions is a fundamental technique.
Machine learning problems often involve learning distributions
over structured objects such as sentences, images, audio, biological sequences,
and source code. 
These distributions usually factorize into a product of 
conditional distributions, e.g., the probability of a word given the previous words. Such distributions are naturally represented as programs
with sampling operations, which we call \emph{randomized programs}.

To predict a structured output,
traditional search methods like beam search compute a set of predictions that 
approximately maximize the probability according to a learned model.
However, recent work has instead favored sampling from the model, with the goal of obtaining diverse and higher-quality predictions
\citep{shi19frangel,fan18hierarchical,sbs,holtzman18learning,holtzman19curious,radford19language}.
The same considerations suggest that sampling without replacement could be even more desirable by avoiding duplicate samples, especially in machine learning where the training objective often pushes the output distribution to become extremely peaked.

We present a data structure called \emph{UniqueRandomizer} for sampling without replacement
from randomized programs.
The main advantage is that it is incremental,
allowing for increased flexibility in stopping conditions for the sampling procedure. For instance, one can draw more samples until a solution is found, or until a sample-based estimate has converged, or enough sample diversity is obtained, and so on.
A closely related algorithm for sampling without replacement is Stochastic Beam Search (SBS)~\cite{sbs}, based on the Gumbel-top-$k$ trick.
SBS is mathematically more complex than \emph{UniqueRandomizer} and does not readily support incrementality, but it has advantages in parallelism and in providing an importance-sampling method for statistical estimation.
Fortunately, we are able to present a combined method with both advantages, i.e., the flexibility of incremental sampling and efficiency of batched computations.
We also derive an improved estimator that can be used with any method of sampling without replacement.

We experimentally demonstrate that using \emph{UniqueRandomizer} leads to higher-quality samples in program synthesis and combinatorial optimization.
More broadly, a major contribution of this paper is making the case that \emph{UniqueRandomizer} is a general technique that can be applied to many scenarios.

\section{Approach}
\label{sec:approach}

As a motivating example, consider a 
program synthesis task: given pseudocode for a short program and some input/output examples, generate the corresponding source code. Suppose that we train a neural sequence-to-sequence model 
that takes pseudocode as input and generates candidate programs. 
We would like to sample from the model until we find a program that satisfies the examples.

We do not know upfront how many samples are needed to find a solution, so to minimize computational waste, we want to sample from the model \emph{incrementally}, one sample at a time until a solution is found.
Furthermore, we have no use for duplicate samples because the quality of a sampled program is deterministic. Since we want to trust the trained model's decisions, we may apply a low temperature when sampling, but this in turn increases the chance of sampling duplicate programs. If we sample \emph{without replacement}, then we can obtain higher-quality samples while avoiding duplicates altogether. Rejection sampling is a standard approach for this, but can be inefficient for highly skewed distributions, such as what we expect to obtain from a trained model with low sampling temperature. Our approach solves these issues with an augmented trie, described in Section~\ref{sec:approach-core}.

\begin{figure}
    \begin{lstlisting}[mathescape=true]
def $\calP$([h, W0, W1], $\calC$):
  tokens = []
  for i in range(0, 100):
    h = softmax(matmul(W0, h))
    probs = softmax(matmul(W1, h))
    tokens.append($\calC$(probs))
  return tokens\end{lstlisting}
  \vspace{-8pt}
    \caption{A simple example of a randomized program.
This program
samples from a recurrent neural network
with weights \code{W0} and \code{W1} and initial state \code{h}.
The  random choice operator $\calC$ is used to sample tokens from the outptut distribution \code{probs} at each step of the RNN.}
    \label{fig:example-program-rnn}
\end{figure}

\subsection{Problem Formalization}
\label{sec:approach-formal}

In this section, we formalize the sampling problem
that we consider. To unify the broad class
of distributions that our method applies to, it is convenient 
to represent the distribution of interest as a program.
Specifically, suppose we have a program $\calP$
that defines a function mapping objects of 
an arbitrary type $X$ to those of type $Y$. We further endow
$\calP$ with a second argument that acts as a source
of randomness---a random choice operation $\mathcal{C}(\pi)$, which samples
from the discrete\footnote{A discrete randomized program $\calP$ cannot draw a random floating-point number, such as $\mathit{Uniform}(0, 1)$, because this is not a \emph{discrete} probability distribution. However, the boolean expression \code{Uniform(0, 1) < 0.3} can be rewritten as \code{$\mathcal{C}$([0.3, 0.7]) == 0}, which is allowed in $\calP$.} probability distribution $\pi$, returning a choice $c \in \{0, \dots, \text{len}(\pi) - 1\}$ with probability $\pi_c$, like \code{numpy.random.choice} in Python.
Except for choices produced by $\mathcal{C}$,
the program $\calP$ is deterministic. Note that $\calP$ may take other input arguments, and it can use control flow including loops, conditionals, and recursion.

We call any program  that has this form a
\emph{discrete randomized program}.
This class of programs does not seem to have
a standard name in the literature, but it is quite broad, including 
probabilistic grammars, neural sequence models, and graphical models.
Randomized programs are essentially the subset of probabilistic programs~\citep{probabilistic_programming} without a conditioning operator.
An example of a discrete randomized program is shown in
Figure~\ref{fig:example-program-rnn}, which
samples from a recurrent neural network.
Of course, randomized programs can be more complex than this---for example, the length of the output
can be random, such as for programs that
sample from a probabilistic context free grammar.

A call to a randomized program
defines a distribution $P(y = \calP(x, \calC))$
over outputs $y$.
We assume that the function call $\calP(x, \calC)$ terminates with probability 1,
although in general this can be tricky to ensure \cite{booth1973applying}.
Our goal is to obtain samples $y_1, \ldots, y_N$ from the distribution $P(y)$ incrementally, and without replacement (WOR).
Sampling without replacement can be formalized
as sampling from a sequence of modified distributions
\begin{equation}
P_\WOR(y_i \mid y_{1:i-1}) 
=  P(y_i = \calP(x, \calC) \mid y_i
\not\in y_{1:i-1}).
\end{equation}
By sampling incrementally, we mean that samples $y_i$ are drawn one by one with a minimal amount of computation performed for each sample.
Given previous samples $y_1, \ldots, y_N$ drawn WOR from $\calP$, we may easily obtain a new sample $y_{N+1}$ drawn from $P_\WOR(y_{N+1} \mid y_{1:N})$, without slowing down the sampler as $N$ increases (as is possible in rejection sampling).

\subsection{Sampling WOR with \emph{UniqueRandomizer}}

Our method is able to sample without replacement, even
without modifying the program $\calP$.
We introduce a data structure called \emph{UniqueRandomizer} that defines a drop-in replacement for 
the random choice operator $\mathcal{C}$,
which efficiently keeps track of the samples
made so far to prevent duplicates.
Here, we give an overview
of the method,
while in the next section we describe
the specifics of the \emph{UniqueRandomizer} data structure and its random choice operator.

To do this,
we define two additional concepts.
An execution of $\calP$ produces
a sequence of calls to $\mathcal{C}$. Each such call takes as input a probability distribution $\pi_i$
and outputs a random choice $c_i \in \{0, \ldots, \text{len}(\pi_i) - 1\}$.
We define a \emph{trace} as the sequence of all random choices
$t = [c_1, \dots, c_h]$ produced during a complete execution of $\calP$.
Note that $\calP$ defines a distribution over its traces
\begin{equation}
    \label{eqn:factorized}
    P(t) = \prod_{i=1}^h P(c_i \mid c_1, \dots, c_{i-1}) = \prod_{i=1}^h (\pi_i)_{c_i}.
\end{equation}
\emph{UniqueRandomizer} samples traces of $\calP$, incrementally and without replacement, according to $P(t)$. 

Sampling WOR from traces yields a sample WOR of program
outputs under a particular condition.
We say that $\calP$ is \emph{trace-injective} if $\calP$ necessarily produces different outputs under different traces. This usually occurs when every call to $\calC$ produces a part of the output, such as when sampling from a sequence model.
This seems to be the most common situation
in machine learning, e.g., the RNN example in Figure~\ref{fig:example-program-rnn} is trace-injective.
If $\calP$ is trace-injective, then sampling traces WOR is equivalent to sampling outputs
WOR (which follows from the change-of-variable
rules for discrete distributions), so \emph{UniqueRandomizer} will produce a sample of outputs without replacement.

Trace-injectivity can be characterized more precisely. 
First, we define
a mapping between
traces and program
outputs. Every execution of $\calP$ produces
a trace $t$ and an output $y;$
let $f(t) = y$ where $y$ is the output when $\calP$ executes with trace $t$.
Trace-injectivity means that the map $f$ is injective.
We can extend this map
to trace prefixes by defining
$F(t') = \{ f(t) \mid \text{$t'$ is a prefix of $t$} \}.$ Then the following theorem
says that $\calP$ is trace-injective when
every choice contributes to the final output,
in a certain sense (proof in Appendix~\ref{app:partition}):

\begin{theorem}
$\calP$ is trace-injective $\iff$ for all trace prefixes
$t' = [c_1, \ldots, c_h]$, the set of possible outputs $F(t')$ is partitioned by the next choice $c_{h+1}\sim\calC(\pi_{h+1})$, i.e., the set
$\{ F([c_1, \ldots, c_h, c_{h+1}]) \mid c_{h+1} \in \{0, \dots, \text{len}(\pi_{h+1}) - 1\}\} $ is a partition of $F(t').$
\end{theorem}

\emph{UniqueRandomizer} can be used to sample without replacement from a trace-injective discrete randomized program $\calP$, as shown in Algorithm~\ref{alg:sample_wor}.
To obtain $k$ samples, we simply run $\calP$
for $k$ iterations, providing it with \emph{UniqueRandomizer}'s
\textsc{RandomChoice} function that remembers
the sequence of choices that are made by each
invocation of $\calP$, and prevents duplicate traces
from being generated.
The way in which we do this,
as well as the implementation of the functions \textsc{Initialize}, \textsc{RandomChoice}, and  
\textsc{ProcessTermination}, are described in the next section.

\begin{algorithm}
\caption{Using \emph{UniqueRandomizer} to sample outputs of $\calP$ without replacement.}
\label{alg:sample_wor}
\begin{algorithmic}[1]
\Procedure{SampleWor}{$\calP, x, k$} 
  \State $\mathit{samples} \gets $ [] 
  \State \Call{Initialize}{$ $}
  \For{$i \in \{1, 2, \ldots, k\}$} 
    \State $y \gets \calP(x, \textsc{RandomChoice})$
    \State $\mathit{samples.append}(y)$ 
    \State \Call{ProcessTermination}{$ $}
  \EndFor
  \State \Return $\mathit{samples}$
\EndProcedure
\end{algorithmic}
\end{algorithm}

\subsection{The \emph{UniqueRandomizer} Data Structure}
\label{sec:approach-core}

\begin{figure*}
\centering
\begin{subfigure}[b]{0.5\textwidth}
    \captionsetup{width=0.9\textwidth}
    \begin{lstlisting}[mathescape=true]
def $\calP$($\calC$):
  length = $\calC$($[0.5, 0.4, 0.1]$)
  sequence = []
  for i in range(0, length):
    sequence.append($\calC$($[0.75, 0.25]$))
  sequence.append($\calC$($[0.1, 0.9]$))
  return sequence\end{lstlisting}
    \caption{A simple randomized program $\mathcal{P}$ that defines a distribution over binary sequences of length 1-3. $\mathcal{P}$ repeatedly calls the random choice function $\calC$, which may be provided by \emph{UniqueRandomizer} (Algorithm~\ref{alg:ur}).}
    \label{fig:example-program}
\end{subfigure}%
\begin{subfigure}[b]{0.5\textwidth}
    \captionsetup{width=0.95\textwidth}
    \centering
    \includegraphics[trim=20 80 20 20,clip,width=0.33\linewidth]{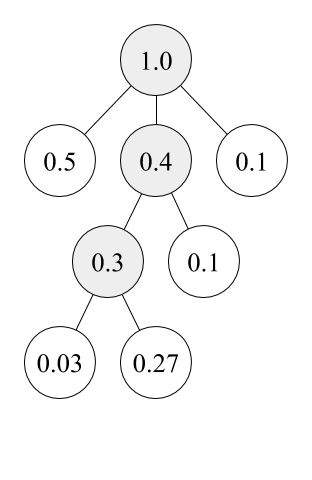}\quad\quad
    \includegraphics[trim=20 80 20 20,clip,width=0.33\linewidth]{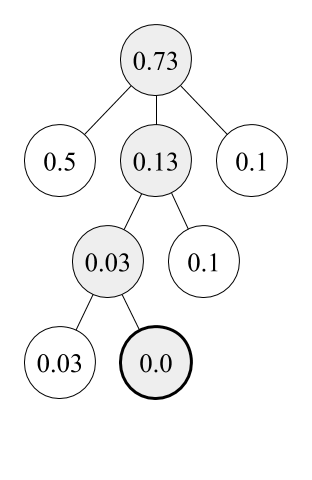}
    \caption{Left: trie after the partial trace $[1, 0]$, immediately before making the third random choice. Right: updated trie after $\mathcal{P}$ terminates for the full trace $[1, 0, 1]$.}
    \label{fig:example-trie}
\end{subfigure}
\caption{An example first run of a simple randomized program $\mathcal{P}$, using \emph{UniqueRandomizer} to sample without replacement (Algorithm~\ref{alg:sample_wor}). Trie nodes store their \emph{unsampled probability mass}. Shaded nodes are those that have been sampled from before, so all of their children are instantiated. Known leaves are outlined in bold. After $\mathcal{P}$ terminates, \textsc{ProcessTermination} is called: the leaf's probability mass of $0.27$ is subtracted from the leaf and its ancestors. On the next run of $\mathcal{P}$, its first random choice request is again $[0.5, 0.4, 0.1]$. However, \emph{UniqueRandomizer}'s \textsc{RandomChoice} function will return an index with probability proportional to the $\mass$ values $[0.5, 0.13, 0.1]$, reflecting the fact that the trace $[1, 0, 1]$ was previously sampled.}
\label{fig:diagram}
\end{figure*}

\begin{algorithm}
    \caption{Random choice operation and
    trie construction for \emph{UniqueRandomizer}.}
    \label{alg:ur}
    \begin{algorithmic}[1]

        \Procedure{Initialize}{$ $}
            \State $\emph{root} \gets $ \Call{TrieNode}{$\emph{parent} = \emptyset, \emph{mass} = 1$}
            \State $\emph{cur} \gets \emph{root}$
        \EndProcedure
        \item[]\vspace{-8pt}
        
        \Procedure{RandomChoice}{$\pi$} 
            \If{\emph{cur}'s children are not initialized yet}
                \For{$0 \le i < \text{len}(\pi)$}
                    \State $\emph{cur}.\emph{children}[i] \gets $ \textsc{TrieNode}(
                    \Statex \hspace{56pt} $\emph{parent} = \emph{cur}, \emph{mass} = \pi[i] \cdot \emph{cur}.\emph{mass})$
                \EndFor
            \EndIf
            \State $\emph{index} \gets $ randomly sample $i$ with probability
            \Statex \hskip\algorithmicindent \phantom{$\emph{index} \gets $} $\propto \emph{cur}.\emph{children}[i].\emph{mass}$
            \State $\emph{cur} \gets \emph{cur}.\emph{children}[\emph{index}]$
            \State \textbf{return} \emph{index}
        \EndProcedure
        \item[]\vspace{-8pt}
        
        \Procedure{ProcessTermination}{$ $}
            \State $\emph{node} \gets \emph{cur}$
            \While{$\emph{node} \ne \emptyset$}
                \State $\emph{node}.\emph{mass} \gets \emph{node}.\emph{mass} - \emph{cur}.\emph{mass}$
                \State $\emph{node} \gets \emph{node}.\emph{parent}$
            \EndWhile
            \State $\emph{cur} \gets \emph{root}$
        \EndProcedure
    \end{algorithmic}
\end{algorithm}

Our main idea is to maintain an augmented trie data structure, which we call the \emph{UniqueRandomizer}, containing the traces that have 
been generated while executing the program $\mathcal{P}$ multiple times.
Nodes in the trie correspond to trace prefixes seen so far. Each edge represents one element of a trace, i.e., a possible outcome for the next call to $\mathcal{C}$. A trie node is a leaf if its trace prefix is actually a full trace, i.e., $\mathcal{P}$ terminates without further calls to $\mathcal{C}$.
 After $\mathcal{P}$ terminates, the trie is updated accordingly.
Figure~\ref{fig:diagram} shows an example.

Each trie node $n$ stores its total \emph{unsampled probability mass}, denoted $\mass(n)$. If $n$ represents the trace prefix $t'$, then $\mass(n)$ equals
\begin{equation}
    \label{eqn:mass}
    \sum_{\text{traces $t$}} \mathds{1}[\text{$t$ is unsampled}] \cdot \mathds{1}[\text{$t'$ is a prefix of $t$}] \cdot P(t).
\end{equation}
We do not compute this sum directly because there will usually be too many traces to enumerate. Instead, we can compute the initial $\mass(n)$ value
for a trie node $n$ using Equation~\eqref{eqn:factorized}, and then update it incrementally after a trace is sampled (when $\mathcal{P}$ terminates). This is shown in Algorithm~\ref{alg:ur}.

First, \textsc{Initialize} is called exactly once, before the program $\calP$ is run. When $\mathcal{P}$ requests a new random choice (\textsc{RandomChoice} in Algorithm~\ref{alg:ur}), we look up the ``current'' trie node $\mathit{cur}$
that corresponds to the state of $\mathcal{P}$'s execution. If $\mathit{cur}$ has not been reached before in a previous execution, we initialize its children. Then, we sample a child $n_i$ of $\mathit{cur}$ with probability proportional to $\mass(n_i)$, update $\mathit{cur}$ to $n_i$, and return $i$.
When one execution of $\mathcal{P}$ terminates, we have sampled a full trace $t_s$, and execute \textsc{ProcessTermination} in Algorithm~\ref{alg:ur}. We mark the corresponding node $n_l$ as a leaf, and we must now update the $\mass$ values. The affected nodes
are $n_l$ and all of its ancestors, 
corresponding to all prefixes of $t_s$, including $t_s$ itself.  For each affected node $n_a$, we update $\mass(n_a) := \mass(n_a) - P(t_s)$, where $P(t_s)$ equals $\mass(n_l)$ before it is updated. Appendix~\ref{app:proof} proves that this scheme results in traces sampled exactly from $P(t)$ without replacement.

Finally, it is useful to detect when all possible traces have been sampled. Mathematically, this simply involves checking if the root node has zero unsampled probability mass. In practice however, accumulation of floating-point errors makes this unreliable. Appendix~\ref{app:exhausted} describes a solution.

\subsection{Extensions and Optimizations}
\label{sec:approach-extensions}

\paragraph{Skipping Probability Computations}
Notice that the probability distributions computed by $\calP$ and passed to $\calC$ are only needed to compute initial $\mass$ values for the corresponding trie node's children. Thus, $\calP$ could be modified to only compute probability distributions when needed, depending on whether the current node's children have already been initialized. This optimization can make sampling with \emph{UniqueRandomizer} even faster than running $\calP$ in a plain loop (i.e., na\"ive sampling with replacement), especially when the same initial program states are seen many times while drawing samples.

\paragraph{Incremental Batched Sampling}
One downside of \emph{UniqueRandomizer} is that runs of $\mathcal{P}$ are difficult to parallelize, since the trie must be updated after each run of $\mathcal{P}$ before the next run can start.
\citet{sbs} previously introduced a different approach to sampling without replacement, called Stochastic Beam Search (SBS), which is a modification of beam search using the Gumbel-top-$k$ trick to sample beam state expansions WOR. Like normal beam search, SBS allows for parallelization when expanding beam states. Section~\ref{sec:related_work} compares \emph{UniqueRandomizer} and SBS in detail.

It is actually possible to combine the strengths of \emph{UniqueRandomizer} and SBS, resulting in a method of incremental batched sampling where each batch allows for parallelization and further batches can be sampled without replacement. Intuitively, one can think of \emph{UniqueRandomizer} as storing the factorized probability distribution of the next sample, conditioned on the fact that previous samples can no longer be chosen. SBS is run with this probability distribution to select the next batch of samples without replacement, where beam states in SBS correspond to \emph{UniqueRandomizer}'s trie nodes. After SBS returns a batch of samples, the $\mass$ values in the \emph{UniqueRandomizer} trie are updated so that the new samples cannot be chosen by later batches.

\paragraph{Locally Modifying Probabilities}
By storing more information in the trie, we can enable efficient local updates to the factorized probability distribution, allowing it to change over time in response to new data while still avoiding previously-seen samples. See Appendix~\ref{app:modify} for details.

\subsection{Estimating Expectations}
\label{sec:approach-estimation}

Many statistical estimators and learning methods expect i.i.d.\ samples;
is there a way to reweight WOR samples so that they can be used
as if they were i.i.d.?
Suppose that we have samples $s_1, \dots, s_k$ drawn without replacement from an arbitrary distribution $p(s)$, perhaps using \emph{UniqueRandomizer} or
some other algorithm,
and we wish to estimate the expectation $\mathbb{E}_{s \sim p}[f(s)]$ for some function $f$.
Estimating an expectation from
WOR samples is a fundamental 
problem in survey statistics \cite{horvitz1952generalization}, but
many of 
these methods do not scale computationally to non-uniform distributions
over large outcome spaces.

\citet{sbs} present a clever solution to this problem.  When asked to produce
$k$ samples WOR from $p$, SBS  
actually produces a sequence
of samples $(s_i, G_i),$
where the set $\{s_1, \ldots, s_k\}$
is the desired WOR sample,
and each $G_i$ is an auxiliary variable drawn from a Gumbel distribution \citep{gumbel}.\footnote{\citet{sbs} and \citet{gumbel_machinery} provide an excellent overview of Gumbels. 
For more on the Gumbel-top-$k$ trick, see \citet{vieira2014efficient} and \citet{sbs}.}
\citeauthor{sbs} use these Gumbels
 to construct an unbiased estimator of the desired expectation,
which we call the \emph{threshold Gumbel estimator (TGE)}:
\begin{equation}
    \label{eqn:estimator}
    \mathbb{E}_p[f(s)] \approx \sum_{i=1}^k w(s_i)f(s_i), \text{ where } w(s_i) = \frac{p(s_i)}{q_\kappa(s_i)}.
\end{equation}
The ``threshold'' $\kappa$ is the $(k+1)$-th largest Gumbel variate obtained during SBS, and $q_\kappa(s_i) = P(\gumbel(\log p(s_i)) > \kappa)$, one minus the Gumbel CDF.
As usual, it is possible to define a variant that normalizes the weights, introducing bias while 
often reducing variance~\cite{sbs}:
\begin{equation}
    \label{eqn:estimator-normalized}
    \mathbb{E}_p[f(s)] \approx \frac{\sum_{i=1}^k w(s_i)f(s_i)}{\sum_{i=1}^k w(s_i)}.
\end{equation}
We derive an equivalent estimator for \emph{UniqueRandomizer} (which, unlike SBS, does not draw Gumbel variates or produce a $\kappa$ threshold). We call this the \emph{Hindsight Gumbel Estimator} (HGE) because we first draw samples and then draw a set of Gumbel variates conditioned on the samples. After drawing the Gumbels ``in hindsight'' to obtain $\kappa$, we directly apply Equation~\eqref{eqn:estimator} or~\eqref{eqn:estimator-normalized}. In fact, this technique applies to any method of sampling without replacement. For example, if the probabilities can be enumerated, one can sample an element according to the given probabilities, set that element's probability to zero, and renormalize the remaining probabilities before drawing the next sample.

To obtain $\kappa$ from samples obtained with \emph{UniqueRandomizer}, we draw a decreasing sequence of $k+1$ Gumbel variates $G_1, \dots, G_{k+1}$ to match the samples $s_1, \dots, s_k$ (and the remaining unsampled probability mass), as if we had used the Gumbel-top-$k$ trick~\cite{vieira2014efficient} with $G_1, \dots, G_{k+1}$ to sample $s_1, \dots, s_k$. In the Gumbel-top-$k$ trick, one first draws a Gumbel variate from $\gumbel(\log p(s))$ for every element $s$ in some sample space $S$. By selecting the maximum $k$ such Gumbels, one actually obtains a WOR sample of $k$ elements from $S$ from the distribution $p$.

The SBS algorithm implicitly defines a joint distribution
$P(G_1, \dots, G_{k+1}, s_1, \dots, s_k)$. From WOR samples $s_1, \dots, s_k$ produced by \emph{UniqueRandomizer}, we sample the hindsight Gumbels from the conditional distribution
$P(G_1, \dots, G_{k+1} \mid s_1, \dots, s_k)$
induced by SBS\footnote{Sampling from this conditional distribution is related to the single-sample retrospective Gumbel question considered by \citet{gumbel_machinery} and \citet{laurent_blog}.}. To do this, we use a key property of Gumbels: if we draw $G_i \sim \gumbel(\log p(s_i))$ for every element $s_i$ in some sample space, then $\max_i G_i \sim \gumbel(\log \sum_i p(s_i))$.

Because $s_1$ is the first sample, $G_1$ is the maximum Gumbel, so we draw $G_1 \sim \gumbel(\log(1))$. Then, we draw the subsequent $G_i$, for $i = 2, \dots, k+1$, in order: at every iteration, the remaining items have a total probability of $1 - \sum_{j=1}^{i-1} p(s_j)$, so we draw
\begin{equation}
    G_i \sim \gumbel\bigg(\log\Big(1 - \sum_{j=1}^{i-1} p(s_j)\Big) \ \Big\vert\ G_{i-1} > G_i\bigg),
\end{equation}
where the condition reflects the fact that $s_{i-1}$ was sampled before $s_i$. Appendix B of \citet{sbs} describes a numerically stable way to draw $G_i$ from this truncated Gumbel distribution. At the end, we assign $\kappa := G_{k+1}$ and apply Equation~\eqref{eqn:estimator} or~\eqref{eqn:estimator-normalized}.
This procedure samples from the same joint
distribution over $G_1, \dots, G_{k+1}, s_1, \dots, s_k$ as SBS. Therefore, HGE is equivalent to TGE
in the sense that the two estimators have the same distribution, but HGE is more generally applicable to any algorithm for sampling without replacement.

Intuitively, the variance in HGE can be attributed to the samples $s_1, \dots, s_k$ and the stochastically-chosen $\kappa$. We can reduce variance in the latter case by repeating the ``hindsight Gumbel'' process to draw multiple $\kappa$ (using the same samples), producing multiple HGE estimates. Averaging these gives the \emph{Repeated HGE} estimate with lower variance.

\section{Analysis and Comparison to Related Work}
\label{sec:related_work}
\paragraph{SBS} Stochastic Beam Search~\cite{sbs} is a prior method of sampling WOR from sequence models. It is similar to \emph{UniqueRandomizer} but has key differences.

First, one main advantage of \emph{UniqueRandomizer} is that it draws samples incrementally, while SBS returns a fixed-size batch of samples. Thus, \emph{UniqueRandomizer} allows for increased flexibility in the number of samples drawn, e.g., drawing samples until a solution is found, until a timeout is reached, after sampling a target fraction of the probability space, until reaching enough diversity in samples, or after the Hindsight Gumbel Estimator (Section~\ref{sec:approach-estimation}) begins to converge. These kinds of stopping criteria would not be possible with SBS\footnote{Although beam search and SBS produce batches of distinct samples, drawing multiple batches may lead to duplicates between batches. One also cannot reuse computation performed for one batch without an auxiliary data structure, which is the purpose of the trie in the incremental batched version of \emph{UniqueRandomizer}.}. Because \emph{UniqueRandomizer} can be used with any arbitrary stopping criterion, it enables sampling-based approaches to be applicable in more scenarios. For example, given a model that generates candidate solutions to a Traveling Salesman Problem instance, SBS could be used to find the best tour among a fixed number of samples, but it would not be able to efficiently find a tour with cost below some threshold. In contrast, \emph{UniqueRandomizer} is applicable in the latter case due to its incrementality.

Second, like normal beam search, the input to SBS
is a next-state function
that enumerates the children of a state. 
For randomized programs $\mathcal{P}$, 
this requires ``pausing'' and ``unpausing'' 
$\mathcal{P}$'s execution.
Thus, the state must contain all of the program context (e.g., local variables) necessary to resume execution.
While this is easy for a sequence model,
it can be challenging for more complex
programs. In contrast, \emph{UniqueRandomizer}
can be elegantly implemented as a wrapper 
on libraries for random number generation,
and it is agnostic to the program context that $\mathcal{P}$ might use.


A third difference is in the number of nodes expanded.
When sampling from machine learning models, it is usually expensive to compute the probability distribution
over a node's children required to expand a node.
In non-degenerate scenarios, \emph{UniqueRandomizer} requires fewer expansions than SBS, since \emph{UniqueRandomizer} only expands the nodes necessary to reach the sampled leaves, while SBS also expands states that later fall off the beam. For example, suppose we sample $k$ sequences of length $L$ with a vocabulary size $V \ge k$. SBS expands exactly $1+(L-1)k$ beam states. 
For  \emph{UniqueRandomizer},
the worst case is when all $k$ sequences have a different first element, so \emph{UniqueRandomizer} expands 
the same number of nodes as SBS.
In the best case, only $L$ expansions are needed if all $k$ sequences differ only at the final position.
We also note that the combination of \emph{UniqueRandomizer} and SBS (incremental batched sampling in Section~\ref{sec:approach-extensions}) provides smooth intermediate behavior: by sampling $b$ batches of size $k/b$, we reduce the number of expansions with high $b$ but obtain better parallelization for low $b$. Setting $b=k$ or $b=1$ reduces to the behavior of \emph{UniqueRandomizer} or SBS, respectively.

Finally, SBS maintains beam nodes for the current and next time step, where each node stores an arbitrarily-complex intermediate state of $\mathcal{P}$. In contrast, \emph{UniqueRandomizer} stores more nodes (i.e., the entire trie), but each node only stores one $\mass$ value. Using \emph{UniqueRandomizer} to sample outputs of $\mathcal{P}$ is at worst a constant factor slower than running $\mathcal{P}$ in a loop\footnote{\textsc{RandomChoice} takes $\Theta(\text{len}(\pi))$ time, but i.i.d.\ sampling requires $\Omega(\text{len}(\pi))$ time anyway to create the distribution $\pi$. \textsc{ProcessTermination} takes $\Theta(h)$ time where $h$ is the leaf depth, which is amortized to an extra $O(1)$ time per random choice.}, but with the optimization in Section~\ref{sec:approach-extensions}, the \emph{UniqueRandomizer} approach may actually be faster.

\emph{UniqueRandomizer} is mathematically simpler than SBS and was developed independently, but we build upon SBS for the incremental batched sampling extension and the Hindsight Gumbel Estimator.

\paragraph{Algorithms for Random Sampling}
\citet{matias2003dynamic} describe a  
dynamic tree data
structure to sample from discrete
distributions with dynamically changing weights, but this is not naturally
adapted to sampling WOR.
Various algorithms
have been proposed for sampling WOR \citep{wong1980efficient,vinterbo2010efficient,Duffield2007priority,efraimidis2006weighted},
but generally these algorithms
do not consider the case of
factorized distributions like \eqref{eqn:factorized}, and so
are inefficient for programs with
potentially long traces.
The relationship between these methods
and the Gumbel-max trick is described by \citet{vieira2014efficient}. The incremental batched version of \emph{UniqueRandomizer}
is related to the top-down Gumbel heap of
\citet{astar_sampling}.
An alternate way of obtaining diverse samples is to explicitly add penalties
to beam search to encourage diversity~\cite{vijayakumar2018diverse,li2016mutual}. It is likely possible that \emph{UniqueRandomizer} can be alternatively implemented using a continuation-passing style and other techniques common in probabilistic programming~\cite{dippl}.

\section{Experiments}

We demonstrate that \emph{UniqueRandomizer} can
lead to improvements in a variety of applications.

\subsection{Program Synthesis}

\begin{figure*}
\centering
\begin{subfigure}[b]{0.34\textwidth}
    \captionsetup{width=0.90\textwidth}
    \includegraphics[height=125pt]{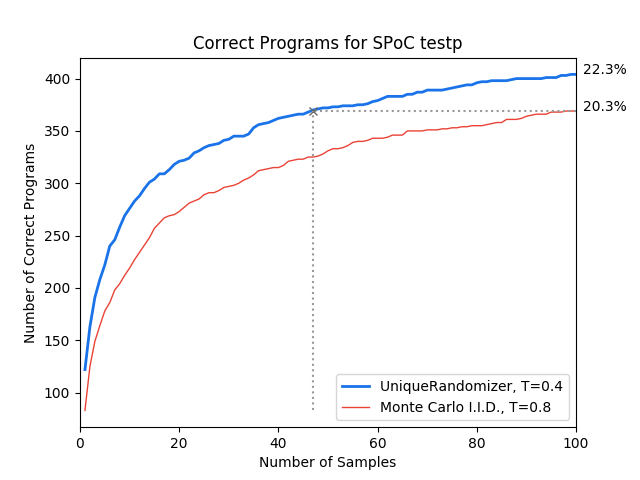}
    \caption{For SPoC testp, in 47 samples, \emph{UniqueRandomizer} achieves the same success rate as i.i.d.\ sampling does in 100 samples.
    }
    \label{fig:spoc-testp}
\end{subfigure}%
\begin{subfigure}[b]{0.34\textwidth}
    \captionsetup{width=0.90\textwidth}
    \centering
    \includegraphics[height=125pt]{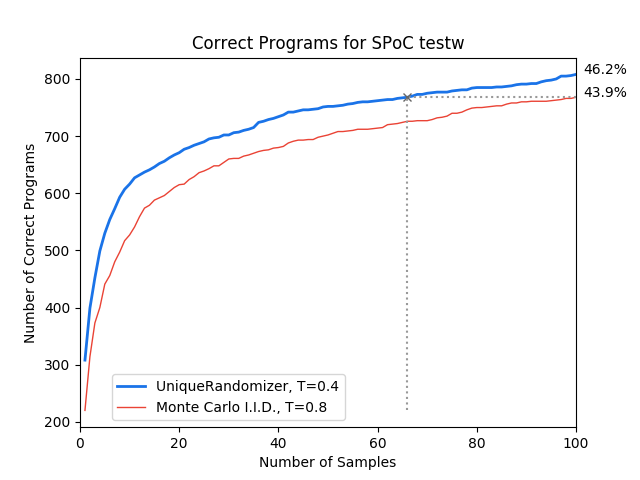}
    \caption{For SPoC testw, in 66 samples, \emph{UniqueRandomizer} achieves the same success rate as i.i.d.\ sampling does in 100 samples.
    }
    \label{fig:spoc-testw}
\end{subfigure}%
\begin{subfigure}[b]{0.32\textwidth}
    \captionsetup{width=0.95\textwidth}
    \centering
    \includegraphics[height=125pt]{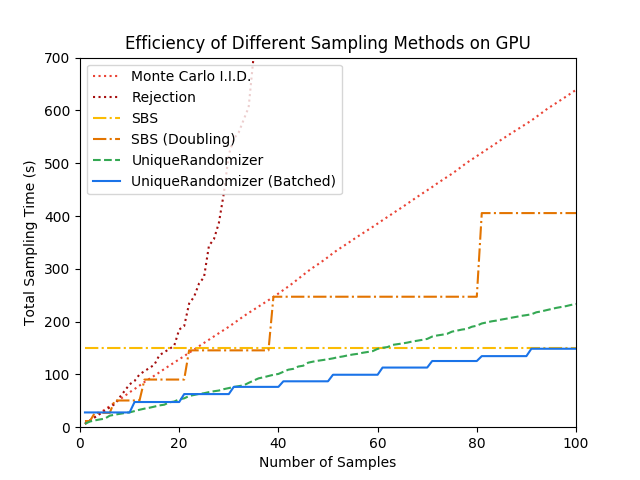}
    \caption{Efficiency of various sampling methods for a 12-line program. The batched \emph{UniqueRandomizer} is the fastest overall.}
    \label{fig:sampling-efficiency}
\end{subfigure}
\caption{Experiment results on the SPoC dataset for program synthesis.}
\label{fig:spoc}
\end{figure*}

We apply \emph{UniqueRandomizer} to program synthesis as described in the motivating example (Section~\ref{sec:approach}).
We use the Search-based Pseudocode to Code (SPoC) dataset~\citep{spoc2019}, which is a collection of 677 problems from programming competitions with 18,356 C++ solution programs written by competitors.
Each line of each program has an accompanying human-authored line of pseudocode. Additionally, each problem has corresponding test cases
that we can use to verify the correctness of a solution program.
The goal is to translate given pseudocode into a code solution for the problem. Compared to natural language translation, the difference is that our ``translation'' is expected to be syntactically correct code consistent with the test cases, and even a single token error likely means that the translation is incorrect.

We train a Transformer model \citep{transformer} on the sequence-to-sequence task of generating a line of code given a line of pseudocode. Details are in Appendix~\ref{app:spoc}. To sample full programs, we iteratively condition the model on each line of pseudocode and sample one code line for each. Concatenating the code lines produces the full program.

We evaluate on the 3,386 pseudocode prompts in the two test splits of the SPoC dataset, ``testp'' and ``testw,'' drawing up to 100 samples for each pseudocode prompt. We compare sampling \emph{with} replacement (using Monte Carlo i.i.d.\ sampling) and sampling \emph{without} replacement (using the batched version of \emph{UniqueRandomizer} with batch size 10). Each sampled program is checked for correctness using the provided test cases.
We use a temperature $\tau=0.8$ when sampling with replacement and $\tau=0.4$ when sampling without replacement, which we found to be the best among $\{0.2, 0.4, 0.6, 0.8, 1.0\}$. Note that a low temperature leads to samples for which the model is more confident but also causes more duplicates when sampling with replacement. 

For both methods of sampling, we report the number
of pseudocode prompts for which a sampled program passes all test cases. Figures~\ref{fig:spoc-testp} and \ref{fig:spoc-testw} show that \emph{UniqueRandomizer}
finds solutions with fewer samples than i.i.d.\ sampling, and succeeds on more prompts overall.

The success rate of our approach is slightly lower than the method in \citet{spoc2019} which achieves 32.5\% success on testp and 51.0\% on testw, with a budget of 100 programs and without using compiler diagnostics for error localization. This may be because \citeauthor{spoc2019} use a more sophisticated model with a copy mechanism and coverage vector, while we simply use a vanilla Transformer. In any case, \citeauthor{spoc2019} find that using compiler diagnostics increases the success rate by 1.7\% on testp and 2.7\% on testw, while our approach of sampling WOR improves over i.i.d.\ sampling by a comparable amount, 2.0\% on testp and 2.3\% on testw.

Our objective in this experiment was not to surpass the state-of-the-art, but rather to show that using \emph{UniqueRandomizer} to draw samples without replacement leads to significant improvement over sampling with replacement. Note that beam search and SBS are existing methods of drawing unique samples, but they are not incremental and would be difficult to use in this setting where one does not know upfront how many samples are needed. Additionally, sampling a program takes about 3 times as long as compiling and executing it on the test cases, so drawing fewer samples is important.

\paragraph{Efficiency}
We also examine the efficiency of various sampling methods in Figure~\ref{fig:sampling-efficiency}, where we draw 100 samples with a GPU using the same medium-sized pseudocode prompt with 12 lines. We observe that \emph{UniqueRandomizer} is more than twice as fast as Monte Carlo i.i.d.\ sampling, explained by the optimization of skipping probability computations (Section~\ref{sec:approach-extensions}). Further note that \emph{UniqueRandomizer} produces unique samples, while Monte Carlo i.i.d.\ sampling does not. Rejection sampling is a na\"ive method of obtaining unique samples but it is slower than Monte Carlo i.i.d\ sampling, in fact becoming progressively slower as more samples are drawn. SBS achieves a low time for 100 samples by using batched computations, but all samples are drawn simultaneously, even if a solution is found within the first few samples, and there is no easy way to draw further unique samples if the first 100 samples are insufficient. The batched version of \emph{UniqueRandomizer} allows for drawing further batches of unique samples, while still taking advantage of batched computation. Thus, if fewer than 100 samples were actually needed, the batched version of \emph{UniqueRandomizer} would be significantly faster than SBS. We also compare to a na\"ive way of using SBS ``incrementally,'' where the batch size doubles on each iteration as $\min\{2 \cdot (\text{previous batch size}), 100\}$ and duplicate samples between batches are discarded, but this method is much slower than \emph{UniqueRandomizer}.

\subsection{Traveling Salesman Problem}

\begin{table*}
    \centering
    \caption{Sampling without replacement from the Attention Model~\citep{kool-tsp} improves upon i.i.d.\ sampling. We show results for 100 and 1280 samples per TSP instance. Concorde~\cite{concorde} is an exact solver, and ``Gap'' is the optimality gap relative to Concorde's exact solution.}
    \setlength{\tabcolsep}{5pt}
    \begin{tabular} {l | c c c | c c c | c c c}
        \toprule
        & \multicolumn{3}{c|}{$n = 20$} & \multicolumn{3}{c|}{$n = 50$} & \multicolumn{3}{c}{$n = 100$} \\
        Method & Cost & Gap & Duplicates & Cost & Gap & Duplicates & Cost & Gap & Duplicates \\
        \midrule
        Concorde (exact) & 3.8357          & 0\%\phantom{.000} &             --                      & 5.696 & 0\%\phantom{.00}         &             --                      & 7.765 & 0\%\phantom{.00}         &             -- \\
        AM, with rep.\ $\times100$       &  3.8397 & 0.105\%          & \phantom{11}96.8                    & 5.735  & 0.69\%         & \phantom{11}63.6                    & 7.979  & 2.77\%         & \phantom{11}4.2 \\
        AM, with rep.\ $\times1280$      &  3.8381 & 0.063\%          &           1274.5                    & 5.724  & 0.49\%         &           1121.3                    & 7.944  & 2.31\%         &           218.9 \\
        AM, w/o rep.\  $\times100$       &  3.8361  & 0.011\%          & \phantom{111}\textbf{0}\phantom{.0} & 5.726  & 0.53\%         & \phantom{111}\textbf{0}\phantom{.0} & 7.979  & 2.76\%         & \phantom{11}\textbf{0}\phantom{.0} \\
\textbf{AM, w/o rep.\ }$\bm{\times1280}$ & \textbf{3.8358} & \textbf{0.002}\% & \phantom{111}\textbf{0}\phantom{.0} & \textbf{5.712} & \textbf{0.29}\% & \phantom{111}\textbf{0}\phantom{.0} & \textbf{7.942} & \textbf{2.28}\% & \phantom{11}\textbf{0}\phantom{.0} \\
        \bottomrule
    \end{tabular}
    \label{tab:tsp-model}
\end{table*}

\begin{table*}
    \centering
    \caption{\emph{UniqueRandomizer} applied to the farthest insertion heuristic for TSP outperforms two of three recent deep-learning approaches~\citep{bello, deudon, kool-tsp}. For methods marked with (*), the results are copied from \citet{kool-tsp}. All of the sampling approaches use 1280 samples per TSP instance.}
    \setlength{\tabcolsep}{8pt}
    \begin{tabular} {l | c c | c c | c c}
        \toprule
        & \multicolumn{2}{c|}{$n = 20$} & \multicolumn{2}{c|}{$n = 50$} & \multicolumn{2}{c}{$n = 100$} \\
        Method & Cost & Gap & Cost & Gap & Cost & Gap \\
        \midrule
        Concorde (exact)                             & 3.8357           & 0\%\phantom{.000} & 5.696           & 0\%\phantom{.00} & 7.765           & \phantom{1}0\%\phantom{.00} \\
        Bello et al., i.i.d.\ sampling (*)           & \multicolumn{2}{c|}{--}              & 5.75\phantom{0} & 0.95\%           & 8.00\phantom{0} & \phantom{1}3.03\% \\
        EAN, i.i.d.\ sampling\ (*)                   & 3.84\phantom{00} & 0.11\%\phantom{0} & 5.77\phantom{0} & 1.28\%           & 8.75\phantom{0} &           12.70\% \\
        \textbf{AM, i.i.d.\ sampling}                & 3.8381           & 0.063\%           & \textbf{5.724}  & \textbf{0.49}\%  & \textbf{7.944}  & \phantom{1}\textbf{2.31}\% \\
        Farthest Insertion, greedy (1 tour)                           & 3.9262           & 2.358\%           & 6.011           & 5.53\%           & 8.354           & \phantom{1}7.59\% \\
        \textbf{Farthest Insertion, \emph{UniqueRandomizer}} & \textbf{3.8372}  & \textbf{0.038}\%  & 5.746           & 0.88\%           & 7.981           & \phantom{1}2.79\% \\
        \bottomrule
    \end{tabular}
    \label{tab:tsp-heuristic}
\end{table*}

Sampling without replacement can be used for combinatorial optimization problems, such as the Traveling Salesman Problem (TSP). Several deep learning models for TSP have been proposed~\citep{bello, deudon, kool-tsp}, where the model outputs candidate tours, and the best tour is chosen among multiple samples drawn from the model. Most recently, the Attention Model by \citet{kool-tsp} was shown to be better than the others on randomly-generated TSP instances with $n = \text{20}$, 50, and 100 nodes.

We compare sampling with and without replacement from the Attention Model, using 1280 samples as in~\citet{kool-tsp}, as well as 100 samples. We also record the number of duplicate tours sampled in each setting, which quantifies the amount of computation wasted by i.i.d.\ sampling (with replacement). Table~\ref{tab:tsp-model} shows the results on \citeauthor{kool-tsp}'s dataset. Observe that i.i.d.\ sampling (used by \citeauthor{kool-tsp}) results in many duplicate tours, especially on the smaller graphs. The duplicate rate also increases with the number of samples, e.g., from 63.6\% duplicate in 100 samples to 87.6\% duplicate in 1280 samples for $n = 50$. In contrast, sampling without replacement avoids duplicates altogether, leading to better final costs. For $n = 20$, sampling 100 tours without replacement outperforms sampling 1280 i.i.d.\ tours with replacement (with optimality gaps of 0.011\% and 0.063\% respectively).

For this application, incremental sampling with \emph{UniqueRandomizer} is 5-25\% faster than na\"ive incremental i.i.d.\ sampling due to the optimization from Section~\ref{sec:approach-extensions}. In the batched case, sampling without replacement using SBS is about 45-80\% slower than batched i.i.d.\ sampling, but for the smaller graph sizes, the slowdown of SBS is outweighed by the benefit of avoiding duplicate samples.

\paragraph{Farthest Insertion} To demonstrate that \emph{UniqueRandomizer} is applicable to a wide range of randomized programs, we also consider the \emph{farthest insertion} heuristic for TSP, which is the best greedy baseline in \citet{kool-tsp}'s results. This heuristic maintains a cycle for a subset of the nodes, and on each iteration, the node that is farthest from the cycle is inserted into the cycle at the cheapest location (such that the new cycle has minimal cost). We transform the greedy heuristic into a discrete randomized program by relaxing the greedy choice for insertion location. Specifically, if inserting a node at location $i$ causes the cycle's cost to increase by $\costDelta(i)$, then we sample an insertion location $i$ with probability proportional to $\costDelta(i)^{-1/\tau}$ where $\tau$ is a temperature hyperparameter. In the limit, $\tau=0$ corresponds to the greedy heuristic, while $\tau=\infty$ corresponds to choosing an insertion location uniformly. We set $\tau=0.3$, $0.2$, and $0.15$ for $n = \text{20}$, 50, and 100 nodes, respectively. These choices of $\tau$ were obtained from a simple search over the range $0.05 \le \tau \le 0.5$.

We then use \emph{UniqueRandomizer} to sample candidate tours without replacement from the modified farthest insertion heuristic. The results are in Table~\ref{tab:tsp-heuristic}. We found that the heuristic approach actually outperforms two of the three learned models included in \citet{kool-tsp}'s comparison, in fact outperforming all three for $n = 20$.

This result is encouraging given the simplicity of this experiment. We started with a well-known greedy heuristic, and by changing only a few lines of code, we turned it into a randomized program via a straightforward relaxation and used \emph{UniqueRandomizer} to sample without replacement. By tuning only one hyperparameter (the temperature $\tau$), and without any training, we obtained results that are competitive with those of carefully-constructed deep learning models.

\subsection{(Repeated) Hindsight Gumbel Estimator}

\begin{figure}[tb!]
    \centering
    \includegraphics[height=150pt]{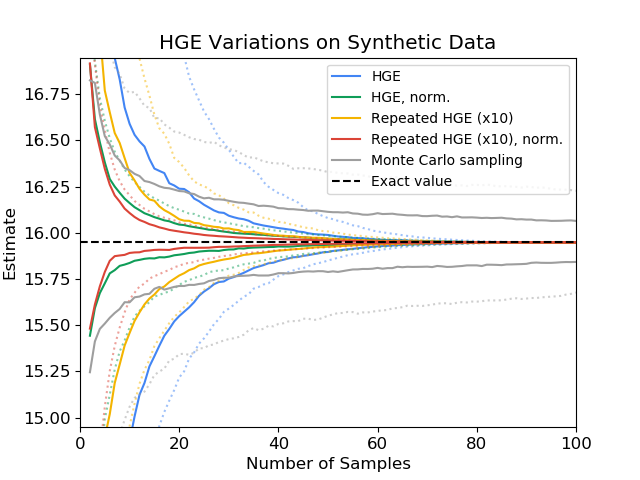}
    \caption{HGE variations on synthetic data. The sample space has 100 elements, and all HGE variations converge to the exact value after all elements are sampled. We drew 2000 sequences of samples. Dotted and solid lines show the inner 90\% and 50\% of the data, respectively.}
    \label{fig:hge}
\end{figure}

In Section~\ref{sec:approach-estimation} we proposed the Hindsight Gumbel Estimator for $\mathbb{E}_p[f(s)]$, the expectation of a function of samples drawn without replacement from a distribution $p$. HGE can be normalized using Equation~\eqref{eqn:estimator-normalized} and/or repeated. Figure~\ref{fig:hge} shows the performance of the estimators on synthetic data. The Monte Carlo estimate simply averages $f(s)$ for i.i.d.\ $s$ sampled with replacement.
We see that HGE, normalized and repeated for 10 iterations, has the lowest variance. We chose a heavily-skewed $p$ so that sampling with replacement encounters many duplicates, and we enforce a strong correlation between $p$ and $f$ (or else incorrect estimators might appear to do well). These properties are common when performing estimation in the context of machine learning.

\section{Conclusion}
We presented \emph{UniqueRandomizer}, an efficient data structure for incremental sampling without replacement. We also derived the Hindsight Gumbel Estimator, a new estimator for samples drawn without replacement that has lower variance than similar previous estimators. Our experiments show that sampling without replacement leads to significant improvements over i.i.d.\ sampling in program synthesis and combinatorial optimization. The incremental nature of \emph{UniqueRandomizer} is especially important in domains like program synthesis where the number of required samples is not known upfront. By eliminating redundant probability computations, \emph{UniqueRandomizer} also allows for very efficient sampling, and the batched version is even more efficient.

Because \emph{UniqueRandomizer} is compatible with a broad class of programs, we believe it could be applicable to other problems beyond those explored in this paper. As with our program synthesis and TSP experiments, \emph{UniqueRandomizer} can be used to sample distinct outputs of a neural model to solve search problems (including constraint satisfaction problems) or obtain diversity in natural language generation~\cite{sbs}. One can also use \emph{UniqueRandomizer} to perform distinct rollouts in reinforcement learning~\cite{kool-buy} or Monte Carlo Tree Search~\cite{mcts} to estimate the value of a state with the Hindsight Gumbel Estimator until that estimate converges sufficiently. \emph{UniqueRandomizer} may also be useful in the context of randomized rounding~\cite{randomized-rounding} and probabilistic programming~\cite{probabilistic_programming}.

Our Python implementations of \emph{UniqueRandomizer}, its batched version, Stochastic Beam Search, and the Hindsight Gumbel Estimator can be found at \url{https://github.com/google-research/unique-randomizer}.

\section*{Acknowledgments}
The authors would like to thank David Dohan, Andreea Gane, David Belanger, Danny Tarlow, and Percy Liang for helpful conversations and explorations.

\bibliographystyle{plainnat}
\bibliography{references}

\clearpage

\begin{appendix}

\section{Proof of Theorem 1}
\label{app:partition}

\setcounter{theorem}{0}
\begin{theorem}
$\calP$ is trace-injective $\iff$ for all trace prefixes
$t' = [c_1, \ldots, c_h]$, the set of possible outputs $F(t')$ is partitioned by the next choice $c_{h+1}\sim\calC(\pi_{h+1})$, i.e., the set
$\{ F([c_1, \ldots, c_h, c_{h+1}]) \mid c_{h+1} \in \{0, \dots, \text{len}(\pi_{h+1}) - 1\}\} $ is a partition of $F(t').$
\end{theorem}

\begin{proof}
For any trace prefix $[c_1, \ldots, c_h]$,
define the collection
\begin{align*}
    \calF([c_1, \ldots, c_h]) = \{& F([c_1, \ldots, c_h, c_{h+1}]) \\
    & \mid c_{h+1} \in \{0, \dots, \text{len}(\pi_{h+1}) - 1\}\}.
\end{align*}

First, if $\calP$ is trace-injective, 
then for any trace prefix $[c_1, \ldots, c_h]$, and any $c \neq c'$,
let $y_1 \in F([c_1, \ldots, c_h, c])$
and $y_2 \in F([c_1, \ldots, c_h, c'])$.
Then there exist traces 
\begin{align*}
    t^{(1)} = [c_1, \ldots, c_h, c^{(1)}_{h+1}, \ldots, c^{(1)}_{h_1}]
\end{align*}
and
\begin{align*}
    t^{(2)} = [c_1, \ldots, c_h, c^{(2)}_{h+1}, \ldots, c^{(2)}_{h_2}]
\end{align*}
such that 
\begin{enumerate}[label=(\alph*)]
    \item $c^{(1)}_{h+1} = c$ and $c^{(2)}_{h+1} = c'$,
    \item $f(t^{(1)}) = y_1$ and $f(t^{(2)}) = y_2$.
\end{enumerate}
Clearly $t^{(1)} \neq t^{(2)},$ so because $f$ is injective, we have that $y_1 \neq y_2$.
Also $\cup_{c'} F([c_1, \ldots, c_h, c']) = F([c_1, \ldots, c_h])$.
This means that
$\calF([c_1, \ldots, c_h])$ is a partition
of $F([c_1, \ldots, c_h])$.

Conversely, assume that $\calF(t')$ is a partition of $F(t')$ for any trace prefix $t'$. Let
\begin{align*}
    t^{(1)} = [c^{(1)}_1, \ldots, c^{(1)}_{h_1}]
\end{align*}
and
\begin{align*}
    t^{(2)} = [c^{(2)}_1, \ldots, c^{(2)}_{h_2}]
\end{align*}
be distinct traces. Let $h$ be the length of their longest common prefix, so $c^{(1)}_i = c^{(2)}_i$ for all $1 \le i \le h$, and $c^{(1)}_{h+1} \ne c^{(2)}_{h+1}$. By definition,
\begin{align*}
    f(t^{(1)}) \in F([c^{(1)}_1, \ldots, c^{(1)}_h, c^{(1)}_{h+1}])
\end{align*}
and
\begin{align*}
    f(t^{(2)}) &\in F([c^{(2)}_1, \ldots, c^{(2)}_h, c^{(2)}_{h+1}]) \\
    &= F([c^{(1)}_1, \ldots, c^{(1)}_h, c^{(2)}_{h+1}]).
\end{align*}
But these two sets are disjoint,
because $\calF([c^{(1)}_1, \ldots, c^{(1)}_h])$ partitions the set
$F([c^{(1)}_1, \ldots, c^{(1)}_h])$.
Therefore, $f(t^{(1)}) \neq f(t^{(2)})$, establishing that $f$ is injective and that $\calP$ is trace-injective.
\end{proof}

\section{Proof of Correctness}
\label{app:proof}

\begin{theorem}
    Let $\mathcal{P}$ be a discrete randomized program that terminates, and let $P(t)$ be the probability that $\mathcal{P}$ runs with trace $t$. Suppose we have already sampled distinct traces $t_1, \dots, t_j$. If, at any \emph{UniqueRandomizer} trie node $n$ we move to a child $c$ with probability proportional to $\mass(c)$, then upon reaching a leaf node, the resulting trace is drawn from $P(t \mid t \not \in \{t_1, \dots, t_j\})$.
\end{theorem}
\begin{proof}
    Let $n_0, \dots, n_h$ be any root-to-leaf path, where $n_0$ is the root and $n_h$ is the leaf. Let $t$ be the trace corresponding to $n_h$. According to Equation~\eqref{eqn:mass},
    \begin{equation*}
        \mass(n_h) =
        \begin{cases} 
            0 & \quad \text{if } t \in \{t_1, \dots, t_j\} \\
            P(t) & \quad \text{otherwise.}
        \end{cases}
    \end{equation*}
    We complete the proof by showing that the leaf $n_h$ is reached with the desired probability:
    \begin{align}
        &\phantom{{}={}}P(\text{$n_h$ is reached}) \nonumber \\
        &= \prod_{i=1}^h P(\text{$n_i$ is the selected child of $n_{i-1}$}) \nonumber \\
        &= \prod_{i=1}^h \frac{\mass(n_i)}{\sum_{c \in \children(n_{i-1})} \mass(c)} \nonumber \\
        &= \prod_{i=1}^h \frac{\mass(n_i)}{\mass(n_{i-1})} \label{eqn:proof-part} \\
        &= \frac{\mass(n_h)}{\mass(n_0)} \nonumber \\
        &= \frac{\mass(n_h)}{1 - \sum_{i=1}^j P(t_i)} \nonumber \\
        &= \begin{cases}
                0 & \quad \text{if } t \in \{t_1, \dots, t_j\} \\
                \frac{1}{1 - \sum_{i=1}^j P(t_i)}P(t) & \quad \text{otherwise}
            \end{cases} \nonumber \\
        &= P(t \mid t \not \in \{t_1, \dots, t_j\}). \nonumber
    \end{align}
    Equality \eqref{eqn:proof-part} holds because a non-leaf node's $\mass$ equals the sum of its children's $\mass$ values. 
\end{proof}

\section{Detecting Exhausted Nodes}
\label{app:exhausted}
We say that a trie node is \emph{exhausted} if it has zero unsampled probability mass, i.e., all of its probability mass is sampled. Due to floating-point errors, a node's $\mass$ might not be set to exactly zero after it should be exhausted. We handle this by carefully propagating the information that a given node has zero unsampled probability mass.

When a node $n$ is marked as a leaf, we directly assign $\mass(n) := 0$. Then, when subtracting mass from one of $n$'s ancestors $a$, we first check if $\mass(c) = 0$ for all children $c$ of $a$. If so, we directly set $\mass(a) := 0$ instead of using a subtraction operation. With this approach, a node's $\mass$ will be exactly $0$ after all of its descendent leaves are sampled. Algorithm~\ref{alg:ur-exhausted} includes this process, elaborating on the pseudocode in Algorithm~\ref{alg:ur}.

\begin{algorithm}
    \caption{\emph{UniqueRandomizer}, with careful detection of exhausted nodes}
    \label{alg:ur-exhausted}
    \begin{algorithmic}[1]

        \LineComment{Called once to initialize the data structure}
        \Procedure{Initialize}{$ $}
            \State $\emph{root} \gets $ \Call{TrieNode}{$\emph{parent} = \emptyset, \emph{mass} = 1$}
            \State $\emph{cur} \gets \emph{root}$
        \EndProcedure
        \item[]\vspace{-4pt}
        
        \LineComment{Whether \emph{node} is completely sampled}
        \Procedure{Exhausted}{\emph{node}}
            \If{\emph{node} is a leaf}
                \State \textbf{return} True
            \EndIf
            \If{\emph{node} has never been sampled from before}
                \State \textbf{return} False
            \EndIf
            \State \textbf{return} whether all of \emph{node}'s children have $0 \mass$
        \EndProcedure
        \item[]

        \LineComment{Called when $\mathcal{P}$ requests a random choice}
        \Procedure{RandomChoice}{$\pi$}
            \If{\Call{Exhausted}{$\emph{cur}$}}
                \State \textbf{raise} Error(``no more unique traces exist'')
            \EndIf
            \If{\emph{cur}'s children are not initialized yet}
                \For{$0 \le i < \text{len}(\pi)$}
                    \State $\emph{cur}.\emph{children}[i] \gets $ \textsc{TrieNode}(
                    \Statex \hspace{56pt} $\emph{parent} = \emph{cur}, \emph{mass} = \pi[i] \cdot \emph{cur}.\emph{mass})$
                \EndFor
            \EndIf
            \State $\emph{index} \gets $ randomly sample $i$ with probability
            \Statex \hskip\algorithmicindent \phantom{$\emph{index} \gets $} $\propto \emph{cur}.\emph{children}[i].\emph{mass}$
            \State $\emph{cur} \gets \emph{cur}.\emph{children}[\emph{index}]$
            \State \textbf{return} \emph{index}
        \EndProcedure
        \item[]\vspace{-4pt}
    
        \LineComment{Called after $\mathcal{P}$ terminates}
        \Procedure{ProcessTermination}{$ $}
            \State mark \emph{cur} as a leaf
            \State $\emph{node} \gets \emph{cur}$
            \While{$\emph{node} \ne \emptyset$}
                \If{\Call{Exhausted}{\emph{node}}}
                    \State $\emph{node}.\emph{mass} \gets 0$
                \Else
                    \State $\emph{node}.\emph{mass} \gets \max\{\emph{node}.\emph{mass} - \emph{cur}.\emph{mass},$
                    \Statex \hskip\algorithmicindent\hskip\algorithmicindent\hskip\algorithmicindent \phantom{$\emph{node}.\emph{mass} \gets \max($}$0\}$
                \EndIf
                \State $\emph{node} \gets \emph{node}.\emph{parent}$
            \EndWhile
            \State $\emph{cur} \gets \emph{root}$
        \EndProcedure
    \end{algorithmic}
\end{algorithm}

\section{Locally Modifying the Factorized Probability Distribution}
\label{app:modify}

A slight modification of \emph{UniqueRandomizer}'s trie allows for efficient local updates to the factorized probability distribution. Instead of storing unsampled probability masses of nodes, the modified trie nodes now store the \emph{unsampled fraction} of the node's total probability mass. Edges in the trie now store the initial probability of following that edge from the source node, as given in the probability distribution provided by $\mathcal{P}$.

Note that the unsampled probability mass of a node $n$ is equal to the product of the edge probabilities from the root to $n$, times the unsampled fraction at $n$. Therefore, by accumulating the product of edge probabilities while walking down the trie, we can compute the unsampled probability mass of nodes, so we can recreate the original \emph{UniqueRandomizer} behavior with the modified trie.

This decomposition enables local modifications to the factorized probability distribution. More precisely, suppose that a trie node $n$ has $k$ children, denoted $n_1, \dots, n_k$, and $n$ initially has outward edge probabilities of $p_1, \dots, p_k$. We wish to change these edge probabilities to $p_1', \dots, p_k'$, so that further samples come from the updated probability distribution and previously-seen samples are still avoided. We do this by updating the trie in the following way. First, we directly replace $n$'s outward edge probabilities with the desired $p_1', \dots, p_k'$. Then, we compute the new unsampled fraction at $n$ with a weighted average of $n$'s children:
\begin{align*}
    &\unsampledFraction(n) := \\
    &\sum_{i = 1}^k \edgeProbability(n, n_i) \cdot \unsampledFraction(n_i).
\end{align*}
Finally, we perform a similar update for all of $n$'s ancestors in upward order (with the root being updated last). After these updates, all values in the trie reflect the new probability distribution.

\section{Program Synthesis Experiment Details}
\label{app:spoc}

For the program synthesis task, we train a Transformer model~\citep{transformer}
to translate lines of pseudocode to lines of C++ code.
We use the Transformer implementation in the Trax framework\footnote{\url{https://github.com/google/trax}.}.
The Transformer uses 2 attention heads, 3 hidden layers, a filter size of 1024, and a hidden dimension size of 512.
We train using ADAM with learning rate 0.05 and batch size 512 for 12,000 steps, which is approximately when the models achieve their lowest evaluation loss. We use linear learning rate warmup for the first 1,000 steps.
These hyperparameters were chosen from the search space in Table~\ref{tab:spoc},
selecting the run with the lowest evaluation loss at the end of training. 
As in \citet{spoc2019}, we withhold 10\% of the training examples as the validation set.

Some of the shorter lines of code in the SPoC dataset have no pseudocode. In some of these instances, we augment the line with pseudocode ourselves. Specifically, if the line is exactly ``\code{\}}'' or ``\code{\};}'' we provide pseudocode ``end'',
if the line is exactly ``\code{int main() \{}'' we provide pseudocode ``main'', and if the line is exactly ``\code{return 0;}'' we provide pseudocode ``return''.

  \begin{table*}[]
    \centering
    \caption{The search space used for tuning the Transformer model.}
    \label{tab:spoc}

\begin{tabular}{l | c c}
        \toprule
\multicolumn{1}{c|}{Hyperparameter}    & Search space            & Selected value \\
        \midrule
Learning rate          & \{0.05, 0.075, 0.1, 0.15\}             &  0.05       \\
Hidden layers          & \{1, 2, 3\}                     &  3  \\
Hidden dimension size            & \{512, 1024\} &  512          \\
Attention heads              & \{2, 4\}                  &  2    \\
Filter size            & \{512, 1024\}           &  1024         \\
        \bottomrule
\end{tabular}
\end{table*}

\end{appendix}

\end{document}